# Classifying Single-Trial EEG during Motor Imagery with a Small Training Set

**Author names and affiliations:**

Yijun Wang

Swartz Center for Computational Neuroscience, Institute for Neural Computation, University of California, San Diego, La Jolla, CA92093 USA

Postal address:

UC San Diego, SCCN

9500 Gilman Drive # 0559

La Jolla CA 92093-0559

Email address:

yijun@sccn.ucsd.edu

Phone: (+1)858-822-7550

Fax:     (+1)858-822-7556

**Abstract**

Before the operation of a motor imagery based brain-computer interface (BCI) adopting machine learning techniques, a cumbersome training procedure is unavoidable. The development of a practical BCI posed the challenge of classifying single-trial EEG with a small training set. In this letter, we addressed this problem by employing a series of signal processing and machine learning approaches to alleviate overfitting and obtained test accuracy similar to training accuracy on the datasets from BCI Competition III and our own experiments.

**Index Terms**—Brain-computer interface, motor imagery, small training set, overfitting

## I. INTRODUCTION

In recent years, BCI systems based on classifying single trial EEGs during motor imagery have developed rapidly [1], [2]. The physiological studies on motor imagery indicate that the spatial distribution of EEG differs between different imagined movements, e.g. hand and foot. First, brain activities at mu (8-12Hz) and beta (18-26Hz) rhythms display event-related desynchronization (ERD) in specific scalp areas corresponding to each imagery state [3]. On the other hand, lateral readiness potential (LRP), which is a slowly decreasing potential, can be recorded with the maximum amplitude over the motor cortex contralateral to the involved hand movements, whereas the readiness potential preceding a foot movement shows no lateralization [4].

Machine learning techniques have been successfully applied to BCI research [5]-[7], however, to obtain sufficient training data, a cumbersome calibration measurement must be performed before online applications. Reduction of the training time can make the system more practical, but may aggravate overfitting, which is affected by the size of the training set. In BCI Competition III [2], the dataset IVa posed the challenge of classifying single-trial EEG during motor imagery with a small training set.

The task of machine learning can be formalized as the estimation of a function $f$, which can correctly classify unlabeled samples. The best function $f$ is the one minimizing the expected risk $R[f]$. Commonly, the minimum of the empirical risk $R_{emp}[f]$ is used to approximate the minimum of the expected risk because the expected risk cannot be minimized directly. The empirical risk will converge toward the expected risk when the number of training data is infinitely large. However, with a limited number of training samples, possible large deviation and overfitting might occur. In this instance, a small generalization error cannot be obtained by simply minimizing the training error. One way to avoid the overfitting is to restrict the complexity of the function class *F* that one chooses the function *f* from [8]. A way of controlling the complexity of a function class is given by the Vapnik-Chervonenkis (VC) theory, where the complexity is denoted by the VC dimension. The relationship between the expected risk and the empirical risk can be given by the following inequality:

$$R[f] \leq R_{emp}[f] + \Phi(n/h) \tag{1}$$

where *n* is the number of training samples and *h* is the VC dimension of the function class. $\Phi$ is the risk confidence, which is a decreasing function of *n/h*. When *n* is fixed, a smaller *h* yields a small risk confidence. Therefore, to minimize the generalization error, i.e. the expected risk *R*[*f*], a small training error should be obtained and the VC dimension should be kept as small as possible.

## II. DATA ACQUISITION

Two datasets recorded during motor imagery were analyzed. "Dataset1" was the dataset IVa of BCI Competition III [2], and "dataset2" was from our online BCI experiments. The approaches we proposed here were aimed at dataset1, and then validated by both datasets.

The detailed description of dataset1 can be found in [2]. The single-trial EEGs were recorded during imagination of *right hand* and *right foot* movement without feedback. For each subject, the experiment consisted of four sessions, and each session contained 70 trials (35 trials per class). The proportion of the training set was different for five subjects (80%, 60%, 30%, 20%, and 10% of 280 trials respectively).

Dataset2 were recorded during imagination of *left* and *right* hand movement with visual feedback from five subjects (right-handed, four males and one female, 22-25 years old). The online paradigm of the experiment was described in [9]. With each subject, we recorded four sessions. There were 60 trials in each session (30 trials per class). The first session was taken as the training set (25% of 240 trials).

## III. FEATURE EXTRACTION AND CLASSIFICATION

*A. Feature Extraction*

Three feature extraction methods were presented and compared. The common spatial patterns (CSP) method [5] and the autoregressive model (AR) method [6] were based on ERD of mu and beta rhythms. Another method was to use the mean value of low frequency EEG components on different channels to reflect the spatial distribution of LRP [7], [10]. In this section, all the parameters were specified only for subject *al* in dataset1 and the task was to discriminate imagination states of *foot* and *hand* movements. The method used for parameter selection will be described in Part C.

*1) Common spatial patterns (CSP)*

The main idea of CSP is to use a linear transform, which can maximize the variance of two-class signal matrices, to project the multi-channel EEG data into a low-dimensional spatial subspace. The algorithm is based on the simultaneous diagonalization of the covariance matrices of both classes [5].

The aim of the CSP method was to design two spatial filters ($SF_H$ and $SF_F$), which led to the estimations of task-related source activities ($S_H$ and $S_F$) corresponding to *right hand* and *right foot* respectively. Then, spatial filtering was performed to eliminate the common components and extract the task-related components. $S_H$ and $S_F$ were estimated by $S_H = SF_H \cdot X$ and $S_F = SF_F \cdot X$, where $X$ was a data matrix of preprocessed multi-channel EEG (0.5-4.5s intercepting and 12-14Hz band-pass filtering). After spatial filtering, the feature corresponding to the source activities was defined as:

$$f = [\log(\frac{\text{var}(S_H)}{\text{var}(S_H)+\text{var}(S_F)})]. \tag{2}$$

*2) Autoregressive Models (AR)*

An AR model of order *p* describes the signal *x(n)* in the following form:

$$x(n) = -\sum_{k=1}^{p} a_k x(n-k) + u(n) \tag{3}$$

where, in the ideal case, $u(n)$ is a white noise with variance $\sigma^2$. The AR coefficients can reflect oscillatory properties of the EEG signal, and the variance $\sigma^2$ contains the amplitude information. In our setting, the model order $p$ was selected as 7. The EEG signals were filtered between 8 and 35Hz and the time window is 0.5-4.5s. Common average reference (CAR) [1] was used for high-pass spatial filtering. The channels with the most significant ERD characters were selected for feature extraction. The feature vector was defined as the concatenation of the AR parameters of the selected N channels, i.e.

$$f = [a_{11}a_{12}...a_{1p}\sigma_1^2 \cdots\cdots a_{N1}a_{N2}...a_{Np}\sigma_N^2]. \tag{4}$$

In practice, Fisher criterion was calculated for channel selection. It was applied to each channel separately to get a score showing how informative each channel was with respect to discriminating the two distributions. The channels with the highest scores were selected for classification.

3) *Lateral readiness potentials (LRP)*

The EEG signals were filtered by a zero-phase low-pass filter at 1.5Hz. The data were baseline corrected by subtracting the mean value of the data over the beginning 0.5s segment. The feature based on LRP was defined as the mean value of the data section between $t_1$ and $t_M$ (i.e. sample points from 0.5s to 1.5s). The feature vector consisted of all the features corresponding to the selected N channels determined by Fisher criterion, i.e.

$$f = [f_1 f_2 ... f_N], \quad f_i = \frac{1}{M}(\sum_{k=t_1}^{t_M} x_i(k)) \ i=1,...,N. \tag{5}$$

*B. Selection of feature extraction method*

Because a linear function has a small VC dimension, we used linear discriminant analysis (LDA) to classify the features extracted by different methods. The classifier is defined by a hyper-plane's normal vector *w* and an offset *b* as:

$$y = \varphi(\mathbf{x}) = \text{sign}(\mathbf{w}^T\mathbf{x}+b) \tag{6}$$

where *x* is the feature vector in a high dimensional space. *w* and *b* are determined by Fisher discriminant analysis (FDA) on the training data. Because the VC dimension of LDA in *d* dimensional feature space is *d*+1, the goal is to find the best tradeoff between the training error and the feature dimension. When dealing with few training samples in a high-dimensional feature space (e.g. 118-channel EEG with several features per channel), feature selection is an effective approach to reduce overfitting through limiting the VC dimension of the function class. The non-informative dimensions of the data should be discarded and the features of interest for classification should be retained.

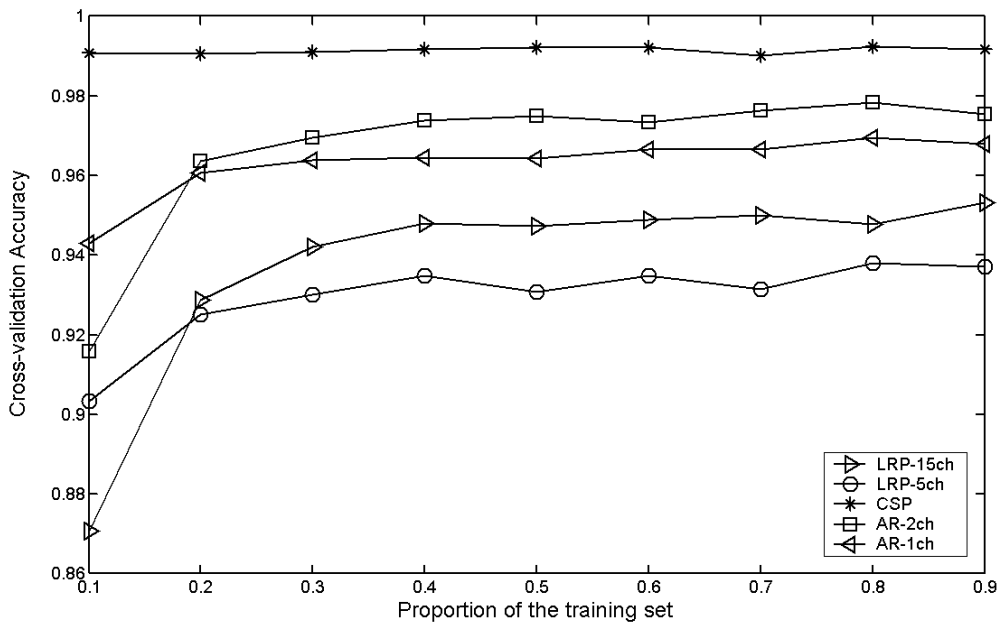

Fig.1 Classification accuracy for subject *al* in dataset1 corresponding to different proportions of the training set when using CSP, LRP, and AR methods to extract the features. LRP-15ch and LRP-5ch indicate that the number of the used channels is 15 and 5 respectively. AR-2ch and AR-1ch are corresponding to 2-channel and 1-channel AR method.

Compared with the other two methods, the CSP method is fit for feature reduction and it can basically reserve the useful information of the multi-channel EEG signals. The number of channels is reduced to a small number through spatial filtering, i.e. the feature space is projected to a low dimension, and therefore, a small risk confidence can be expected. For the other two methods, feature reduction has to be performed by channel reduction. For classification with different size of training set, the number of features in the CSP method can be fixed with the training error slightly changed. But to avoid a large risk confidence caused by the decreasing number of the training samples, the feature space of the AR and RP methods need to be reduced. The cost is that many channels with effective information are discarded and the training error is increased. Fig.1 displays the classification results corresponding to different numbers of training samples on the training set for subject *al* in dataset1. On one hand, the risk confidence increases when the number of training samples decreases, e.g. the accuracy decreased significantly when using LRP and AR methods with only 10% trials as the training set. On the other hand, channel reduction also leads to an obvious decrease of classification performance when using the AR or LRP method. However, the CSP method achieves high and stable performance, which seems to be unaffected by the size of the training set. The results suggest that the CSP method is better for motor imagery EEG classification with a small training set.

*C. Selection of parameters in feature extraction*

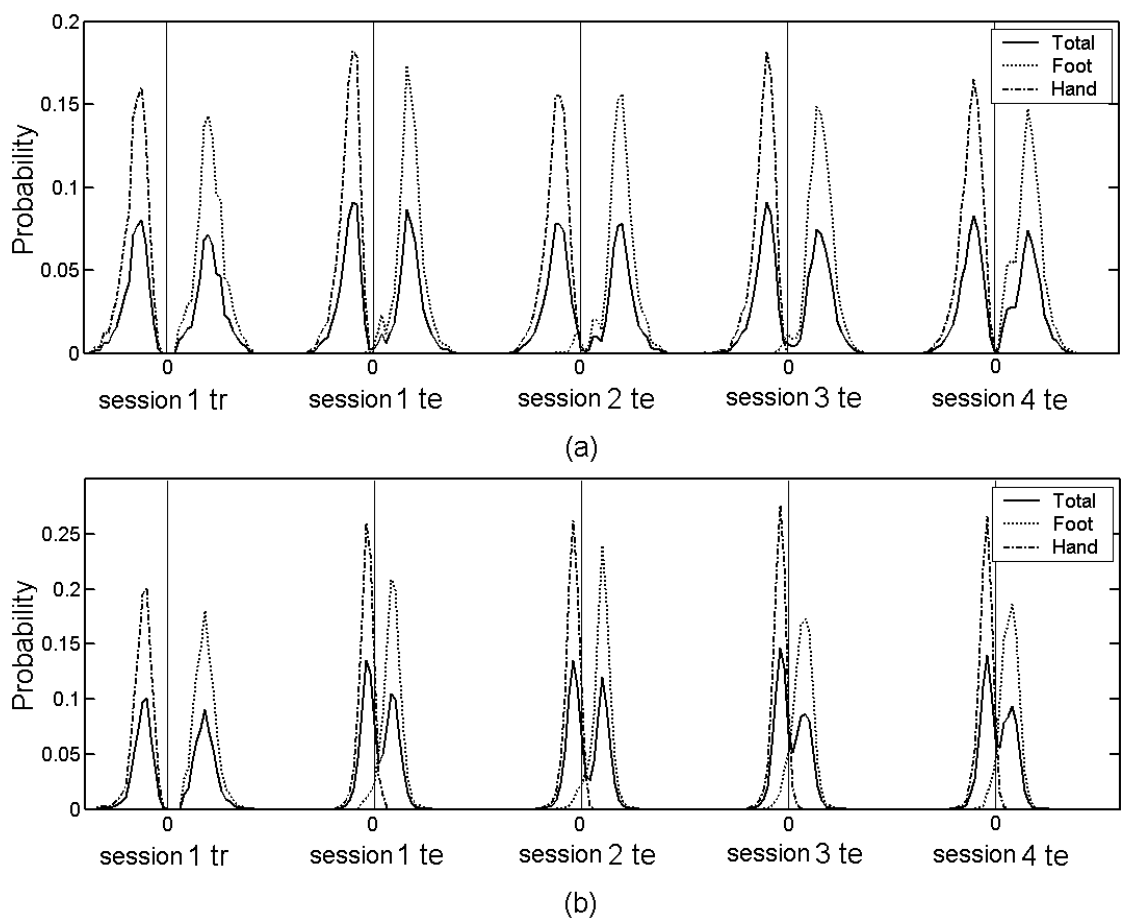

Fig.2 Probability distributions of the LDA output, i.e. *x*, in each session corresponding to (a) good parameters and (b) lower-performance parameters through cross-validations. The elements of *x* were binned into 40 equally spaced containers in (a) and (b) respectively. "tr" and "te" denote the training set and the test set.

Ideally, if the appropriate parameters for data preprocessing in the CSP method, which are commonly derived from a large number of training samples, can be determined with a small training set, a high accuracy can be expected on the test set. Therefore, the challenge was to select the proper parameters with only few training samples. We proposed two criteria for parameter selection. The first criterion is derived from a specific prior knowledge of these datasets that *the numbers of samples for each class are equal*, i.e.

$$P(y=-1) = \int_{-\infty}^{0} pdf(x)dx, \quad P(y=1) = \int_{0}^{+\infty} pdf(x)dx, \quad P(y=-1) = P(y=1) = 0.5 \tag{7}$$

where $x$ is the output of the LDA before applying the signum function and $pdf(x)$ is the probability density function of $x$. $P(y)$ is the probability of class $y$ ("-1" for hand and "1" for foot). This criterion only works when the probability of each class can be estimated. The second is about $pdf(x)$ derived from LDA. Based on the validation by real datasets, we proposed a hypothesis that the extracted feature of the training set follows a distribution similar to the test set. Therefore, *pdf(x) of the test session is similar to pdf(x) of the training session*. The probability distribution with two separate and symmetrical peaks is considered a destination template. Moreover, parameter optimization can be described as maximization of the correlation coefficient between the *pdfs* of the training set and the test set, i.e.

$$\max \rho[pdf(x^{TR}), pdf(x^{TE})] = \frac{\text{cov}[pdf(x^{TR}), pdf(x^{TE})]}{\sigma_{pdf(x^{TR})} \sigma_{pdf(x^{TE})}}. \tag{8}$$

As shown in Fig.2(a), the distributions of all the test sessions are similar to the training set, so a high test accuracy can be expected, whereas the parameters corresponding to Fig. 2(b) still need to be adjusted (The correlation coefficient is 0.81 and 0.04, while the test accuracy is 99.0% and 92.4% respectively). According to the destination, the parameters including *frequency band*, *time window*, and *channel location* should be considered carefully. The physiological knowledge such as the temporal and frequency characteristic of ERD was used to initialize the frequency band and time window, and then a sliding window method was used to make the adjustments. Besides, the channel locations were determined with the aid of the spatial mapping of ERD, which was consistent with the region of the primary sensorimotor cortex.

*D. Classification*

*1) Bootstrap Aggregation (Bagging)*

The final decision of classification was made by a bootstrap aggregation (bagging) method using the Fisher discriminant as the component classifier:

$$y_B = \varphi_B(\mathbf{x}) = E_L[\varphi(\mathbf{x}, L^{(B)})] \tag{9}$$

where $L^{(B)}$ is the bootstrap samples from the training set $L$ and $E_L$ denotes expectation over $L$.

Bagging was introduced by Breigman as a method of reducing the prediction variance without affecting the prediction bias [11]. The basic idea of bagging is to use various subsets of the initial training set to construct a number of component classifiers and the final prediction is determined by a majority vote. The forecasts, given by unstable laws over various subsets, have no deciding vote at the end, while really steady laws have a decisive role. Thus, the predictive performance, as a rule, can be improved. Bagging works well especially when the classifier is unstable, i.e. the predictions are sensitive to small changes in the training data. In our bagging method, the subset was formed by choosing random samples with replacement from the total training data. The volume of the subset was set beforehand (e.g. 50 % of the training data). After the construction of a component classifier, the subset was sent back into the initial training data and the process was repeated for a given number of times (e.g. 50 times as suggested by [11]).

*2) Adaptive classification*

With optimized parameters, the CSP method can be expected to achieve stable performance with only a small training set. Furthermore, if the number of the training samples can be increased, the test performance will be more robust (see (1)). To get more training samples during testing procedure, an adaptive approach based on semi-supervised learning [12], which was realized by adopting the former labeled test samples as extended training samples, was employed. The risk confidence of the latter test samples would be decreased if the former samples were classified correctly. Our investigation showed that an evident performance gain could be achieved by this adaptive approach, especially for the latter sessions. There is an important prerequisite that the distribution of examples, which the unlabeled data will help elucidate, is relevant for the classification problem (i.e. the unlabeled data carry useful information for classification) [12]. Fig. 3 is the flowchart of the adaptive approach, where the classification of the test set was performed session by session. The small training set was supposed to be part of the first session. The procedures are described as follows:

*Step 1*: Label the test samples in session1 based on supervised machine learning. Joint parameter selection and classification were run to optimize the parameters for feature extraction, and the labels of the test samples in session1 were made by voting results of the bagging classifier.

*Step 2*: Label the following sessions through a semi-supervised learning manner. For example, to classify session2, all samples of session1 formed a new training set. In order to obtain better parameters with the aid of new training samples, the parameter selection was performed again. Classification of the last two sessions was done in a similar fashion with an enlarged training set.

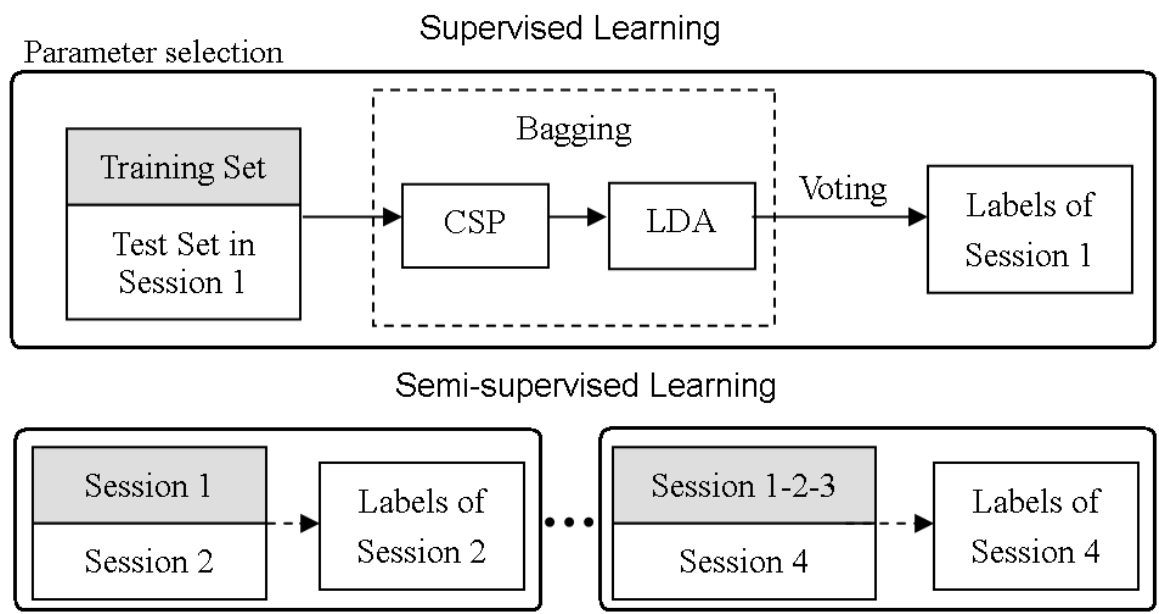

Fig. 3. Flowchart of the adaptive classification. The gray area indicates the traing set.

In this adaptive classification, the bagging method played an important role. It can eliminate the unstable interference caused by the misclassified test samples through voting. With few misclassified test samples, most component classifiers can be considered to be derived from the true-labeled expanded training data. In a simulation testing on subject *al*, we selected 30% of the samples in the first two sessions randomly as the misclassified samples (i.e. labels of these samples were changed manually), and then the last two sessions were used as the test set. The average accuracy of the component classifiers decreased from 99.1% to 83.4 %, however, the accuracy obtained by bagging still retained 95.7% (the volume of the subset was 50% and the number of the component classifier was 50).

## IV. RESULTS AND DISCUSSION

TABLE I

CLASSIFICATION RESULTS OF ALL THE SUBJECTS IN TWO DATASETS

| Subjects | | Training /Test samples | Accuracy ±std (%) | |
|---|---|---|---|---|
| | | | Training Set | Test Set |
| 1 | al | 224/56 | 99.4±0.21 | 100 |
| | aa | 168/112 | 95.1±0.45 | 95.5 |
| | av | 84/196 | 91.3±1.13 | 80.6 |
| | aw | 56/224 | 98.l±1.93 | 100 |
| | ay | 28/252 | 98.2±1.69 | 97.6 |
| 2 | fl | 60/180 | 83.7±6.01 | 86.1 |
| | my | 60/180 | 98.2±1.81 | 100 |
| | sjh | 60/180 | 95.3±2.62 | 99.4 |
| | ww | 60/180 | 98.5±1.80 | 100 |
| | zd | 60/180 | 93.4±3.12 | 95 |
| Mean | | —— | 95.1 | 95.4 |

Table I lists the classification results on the training set (by cross-validation) and the test set. The mean accuracy on the test set was 95.4% with a slight increase comparing with 95.1% of the training set. The result of dataset1 was the winning contribution to BCI Competition III [2]. For subject *aa* and *av*, an approach based on feature combination of CSP, AR and LRP algorithms was used [7]. For other subjects, only the CSP method was used for feature extraction. The approaches we proposed resulted in even higher test accuracy than training accuracy on most subjects (8/10).

Obviously, the overfitting of subject *av* was not entirely avoided. It may attribute to the feature combination algorithm. As we stated above, compared with the CSP method, AR and LRP methods usually lead to a larger risk confidence while the training set is small, so the combination method cannot achieve the performance gain as expected.

Independent from the machine learning algorithm, another approach, which may contribute to solve the problem of a small training set, is to use information from other subjects, i.e. subject-to-subject transfer. So far, no satisfying results with this method have been achieved because the parameters for feature extraction are subject-specific after optimization. Moreover, even for the same subject, a session-to-session transfer is still a challenge due to variant brain states and possible changes caused by the recording equipment, e.g. electrode positions. From this point of view, the combination of the solutions for small training set and session-to-session transfer may be a preferable method, which can make the motor imagery based BCI more practical.

**ACKNOWLEDGMENTS**

The authors are grateful to Klaus-Robert Müller, Benjamin Blankertz and Gabriel Curio for providing their data.

## REFERENCES

[1]  J. R. Wolpaw, N. Birbaumer, D. J. McFarland, G. Pfurtscheller, T. M. Vaughan, "Brain-computer interfaces for communication and control," *Clin. Neurophysiol.*, vol.113, pp.767-791, 2000.

[2]     B. Blankertz, K. -R. Muller, D. J. Krusienski, G. Schalk, J. R. Wolpaw, A. Schlogl, G. Pfurtscheller, Jd. R. Millan, M. Schroder, and N. Birbaumer, "The BCI competition III: validating alternative approaches to actual BCI problems," *IEEE Trans. Rehab. Eng.*, vol.14, no.2, pp.153-159, 2006.

[3]     G. Pfurtscheller, and F. H. Lopes da Silva, "Event-related EEG/MEG synchronization and desynchronization: basic principles," *Clin. Neurophysiol.*, vol.110, pp.1842-1857, 1999.

[4]     K. B. E. Böcker, C. H. M. Brunia, and P. J. M. Cluitmans, "A spatio-temporal dipole model of the readiness potential in humans. II. Foot movement," *Electroencephalogr. Clin. Neurophysiol.*, vol.91, pp.286-294, 1994.

[5]     H. Ramoser, J. M. Gerking, and G. Pfurtscheller, "Optimal spatial filtering of single trial EEG during imagined hand movement," *IEEE Trans. Rehab. Eng.*, vol. 8, no.4, pp.441-446, 2000.

[6]     G. Pfurtscheller, C. Neuper, A. Schlogl, and K. Lugger, "Separability of EEG signals recorded during right and left motor imagery using adaptive autoregressive parameters," *IEEE Trans. Rehab. Eng.*, vol.6, no.3, pp.316-325, 1998.

[7]     G. Dornhege, B. Blankertz, G. Curio, and K. -R. Müller, "Boosting bit rates in noninvasive EEG single-trial classifications by feature combination and multiclass paradigms," *IEEE Trans. Biomed. Eng.*, vol.51, no.6, pp.993-1002, 2004.

[8]     V. N. Vapnik, *The Nature of Statistical Learning Theory*. New York: Springer-Verlag, 1995.

[9]     Y. Wang, B. Hong, X. Gao, and S. Gao, "Phase synchrony measurement in motor cortex for classifying single-trial EEG during motor imagery", Proc. 28th Int. IEEE EMBS Conf., New York, pp. 75-78, 2006.

[10]    Y. Wang, Z. Zhang, Y. Li, X. G, S. G and F. Y, "BCI Competition Data Set IV: an algorithm based on CSSD and FDA for classifying single trial EEG", IEEE Trans. Biomed. Eng., vol.51, no.6, pp.1081-1086, 2004.

[11]    L. Breiman, "Bagging Predictors", *Machine Learning*, vol.24, no.2, pp.123-140, 1996.

[12]    O. Chapelle, B. Schölkopf, and A. Zien, *Semi-supervised Learning*. MIT Press, 2006.